\documentclass[10pt,letterpaper]{article}
\pdfoutput=1

\usepackage[top=0.85in,left=2.75in,footskip=0.75in]{geometry}

\usepackage{amsmath,amssymb}
\usepackage{lingmacros}

\usepackage{changepage}

\usepackage{textcomp,marvosym}

\usepackage{cite}

\usepackage{nameref,hyperref}

\usepackage{xspace}

\usepackage{microtype}
\DisableLigatures[f]{encoding = *, family = * }

\usepackage[table]{xcolor}

\usepackage{array}

\usepackage{placeins}

\usepackage{multirow}

\newcolumntype{+}{!{\vrule width 2pt}}

\newlength\savedwidth

\newcommand\thickhline{\noalign{\global\savedwidth\arrayrulewidth\global\arrayrulewidth 2pt}%
\hline
\noalign{\global\arrayrulewidth\savedwidth}}

\raggedright
\usepackage[aboveskip=1pt,labelfont=bf,labelsep=period,justification=raggedright,singlelinecheck=off]{caption}

\makeatletter
\renewcommand{\@biblabel}[1]{\quad#1.}
\makeatother

\usepackage{lastpage,fancyhdr,graphicx}
\usepackage{epstopdf}
\fancyheadoffset[L]{2.25in}
\fancyfootoffset[L]{2.25in}
\begin{document}
\vspace*{0.2in}

\begin{flushleft}
{\Large
\textbf{\newline{AI, write an essay for me: A large-scale comparison of human-written versus ChatGPT-generated essays}}
}\newline
\\
Steffen Herbold\textsuperscript{1*},
Annette Hautli-Janisz\textsuperscript{1},
Ute Heuer\textsuperscript{1},
Zlata Kikteva\textsuperscript{1},
Alexander Trautsch\textsuperscript{1}
\\
\bigskip
\textbf{1} Faculty of Computer Science and Mathematics, University of Passau, Passau, Germany
\\
\bigskip

* steffen.herbold@uni-passau.de

\end{flushleft}
\section*{Abstract}

\textit{Background:} Recently, ChatGPT and similar generative AI models have attracted hundreds of millions of users and become part of the public discourse. Many believe that such models will disrupt society and will result in a significant change in the education system and information generation in the future. So far, this belief is based on either colloquial evidence or benchmarks from the owners of the models -- both lack scientific rigour. 

\noindent
\textit{Objective:} Through a large-scale study comparing human-written versus ChatGPT-generated argumentative student essays, we systematically assess the quality of the AI-generated content. 

\noindent
\textit{Methods:} A large corpus of essays was rated using standard criteria by a large number of human experts (teachers). We augment the analysis with a consideration of the linguistic characteristics of the generated essays.

\noindent
\textit{Results:} Our results demonstrate that ChatGPT generates essays that are rated higher for quality than human-written essays. The writing style of the AI models exhibits linguistic characteristics that are different from those of the human-written essays, e.g., it is characterized by fewer discourse and epistemic markers, but more nominalizations and greater lexical diversity. 

\noindent
\textit{Conclusions:} Our results clearly demonstrate that models like ChatGPT outperform humans in generating argumentative essays. Since the technology is readily available for anyone to use, educators must act immediately. We must re-invent homework and develop teaching concepts that utilize these AI models in the same way as math utilized the calculator: teach the general concepts first and then use AI tools to free up time for other learning objectives. 



\section*{Introduction}

The massive uptake in the development and deployment of large-scale Natural Language Generation (NLG) systems in recent months has yielded an almost unprecedented worldwide discussion of the future of society. In an open letter at the end of March 2023,\footnote{\url{https://futureoflife.org/open-letter/pause-giant-ai-experiments/}} both influential leaders from the industry like Elon Musk and Steve Wozniak, as well as leading AI experts like Joshua Bengio and Stuart Russel, call for a temporary ban on the development of more powerful AI than GPT-4\cite{openai2023gpt4}. The ChatGPT service which serves as Web front-end to GPT-3.5 \cite{ouyang2022training} and GPT-4 was the fastest-growing service in history to break the 100 million user milestone in January and had 1 billion visits by February 2023.\footnote{\url{https://www.demandsage.com/chatgpt-statistics/}} The call of the open letter is to allow industry and academia to catch up with the recent progress in the field and ``get independent review before starting to train future systems.'' This is the mission of the present paper. 

Driven by the upheaval that is particularly anticipated for education \cite{leahy2023tpack} and knowledge transfer for future generations, we conduct the first independent, systematic study of AI-generated language content that is typically dealt with in high-school education: argumentative essays, i.e., essays in which students discuss a position on a controversial topic by collecting and reflecting on evidence (e.g., `Should students be taught to cooperate or compete?'). While there is a multitude of individual examples and anecdotal evidence for the quality of AI-generated content in this genre,\footnote{e.g., \url{https://www.zdnet.com/article/how-to-use-chatgpt-to-write-an-essay/}} this paper is the first to systematically assess the quality of human-written and AI-generated argumentative texts across different versions of ChatGPT.\footnote{\url{https://chat.openai.com/chat}} We use a fine-grained essay quality scoring rubric based on content and language mastery and employ a  significant pool of domain experts, i.e., high school teachers across disciplines, to perform the evaluation. Using computational linguistic methods and rigorous statistical analysis, we arrive at several key findings: 

\begin{itemize}
    \item AI models generate significantly higher-quality argumentative essays than the users of an essay-writing online forum frequented by German high-school students across all criteria in our scoring rubric.
    \item ChatGPT-4\footnote{ChatGPT web interface with the GPT-4 model.} significantly outperforms ChatGPT-3\footnote{ChatGPT web interface with the GPT-3.5 default model.} with respect to logical structure, language complexity, vocabulary richness and text linking. 
    \item Writing styles between humans and generative AI models differ significantly: for instance, the GPT models use more nominalizations and have higher sentence complexity (signalling more complex, `scientific', language), whereas humans make more use of modal and epistemic constructions (which tend to convey speaker attitude). 
    \item The linguistic diversity of the NLG models seems to be improving over time: while ChatGPT-3 still has a significantly lower linguistic diversity than humans, ChatGPT-4 has a significantly higher diversity than humans. 
\end{itemize}

As such, our work goes significantly beyond existing benchmarks. While OpenAI's technical report on GPT-4 \cite{openai2023gpt4} presents some benchmarks, their evaluation lacks scientific rigour: it fails to provide vital information like the agreement between raters, does not report on details regarding the criteria for assessment or to what extent and how a statistical analysis was conducted for a larger sample of essays. In contrast, our benchmark provides the first (statistically) rigorous and systematic study of essay quality, paired with a computational linguistic analysis of the language employed by humans and two different versions of ChatGPT, offering a glance at how these NLG models develop over time.

\section*{Related work}

\subsection*{Natural Language Generation}

The recent interest in generative AI models can be largely attributed to the public release of ChatGPT, a public interface in form of an interactive chat based on the InstructGPT~\cite{ouyang2022training} model, more commonly referred to as GPT-3.5. In comparison to the original GPT-3~\cite{brown2020language} and other similar generative large language models based on the transformer architecture like GPT-J~\cite{gpt-j}, this model was not trained in a purely self-supervised manner (e.g., through masked language modelling). Instead, a pipeline that involved human-written content was used to fine-tune the model and improve the quality of the outputs to both mitigate biases and safety issues, as well as make the generated text more similar to text written by humans. Such models are referred to as Fine-tune LAnguage Nets (FLANs). For details on their training, we refer to the literature~\cite{wei2022finetuned}. Notably, this process was recently reproduced with publicly available models\footnote{I.e., the complete models can be downloaded and not just accessed through an API.} such as Alpaca~\cite{alpaca} and Dolly.\footnote{\url{https://www.databricks.com/blog/2023/04/12/dolly-first-open-commercially-viable-instruction-tuned-llm}} However, we can only assume that a similar process was used for the training of GPT-4 since the paper by OpenAI does not include any details on model training. Please note that this paper only considers the GPT-4 model without vision, as the latter is not yet publicly available at the date of submission. 

Testing of the language competency of large-scale NLG systems has only recently started. Cai et al. \cite{cai2023does} show that ChatGPT reuses sentence structure, accesses the intended meaning of an ambiguous word, and identifies the thematic structure of a verb and its arguments, replicating human language use. Mahowald \cite{mahowald2023discerning} compares ChatGPT's acceptability judgements to human judgements on the Article + Adjective + Numeral + Noun construction in English. Dentella et al.~\cite{dentella2023testing} show that ChatGPT-3 fails to understand low-frequent grammatical constructions like complex nested hierarchies and self-embeddings. In another recent line of research, the structure of automatically generated language is evaluated.  Guo et al. \cite{guo2023close} show that in question-answer scenarios, ChatGPT-3 uses different linguistic devices than humans. Zhao et al.~\cite{zhao2023chatgpt} show that ChatGPT generates longer and more diverse responses when the user is in an apparently negative emotional state.

Given that we aim to identify certain linguistic characteristics of human-written versus AI-generated content, we also draw on related work in the field of linguistic fingerprinting, which assumes that each human has a unique way of using language to express themselves, i.e., the linguistic means that are employed to communicate thoughts, opinions and ideas differ between humans. That these properties can be identified with computational linguistic means has been showcased across different tasks: the computation of a linguistic fingerprint allows to distinguish authors of literary works \cite{Keim2007-10Liter-5492}, the identification of speaker profiles in large public debates \cite{el-assady-etal-2017-interactive,elassady-etal-20,foulisetal-20a,foulisetal-20b} and the provision of data for forensic voice comparison in broadcast debates \cite{chatzipanagiotidis-etal-2021-broad, ajili-etal-2016-fabiole}. For educational purposes, linguistic features are used to measure essay readability \cite{deutsch-etal-2020-linguistic}, essay cohesion \cite{fiacco-etal-2022-toward} and language performance scores for essay grading \cite{weiss-etal-2019-computationally}. Integrating linguistic fingerprints also yield performance advantages for classification tasks, for instance in predicting user opinion \cite{yang-etal-2020-predicting-personal,tumarada-etal-2021-opinion} and identifying individual users \cite{rocca-yarkoni-2022-language}. 

\subsection*{Limitations of OpenAIs ChatGPT evaluations} 

OpenAI published a discussion of the model's performance of several tasks, including AP classes within the US educational system~\cite{openai2023gpt4}. The subjects used in performance evaluation are diverse and include arts, history, English literature, calculus, statistics, physics, chemistry, economics, and US politics. While the models achieved good or very good marks in most subjects, they did not perform well in English literature. GPT-3.5 also experienced problems with chemistry, macroeconomics, physics, and statistics. While the overall results are impressive, there are several significant issues: firstly, the conflict of interest of the model's owners poses a problem for the performance interpretation. Secondly, there are issues with the soundness of the assessment beyond the conflict of interest, which make the generalizability of the results hard to assess with respect to the models' capability to write essays. Notably, the AP exams combine multiple-choice questions with free-text answers. Only the aggregated scores are publicly available. To the best of our knowledge, neither the generated free-text answers, their overall assessment, nor their assessment given specific criteria from the used judgement rubric are published. Thirdly, while the paper states that 1-2 qualified third-party contractors participated in the rating of the free-text answers, it is unclear how often multiple ratings were generated for the same answer and what was the agreement between them. This lack of information hinders a scientifically sound judgement regarding the capabilities of these models in general, but also specifically for essays. Lastly, the owners of the model conducted their study in a few prompt setting, where they gave the models a very structured template as well as an example of a human-written high-quality essay to guide the generation of the answers. This further fine-tuning of what the models generate could have also influenced the output. The results published by the owners go beyond the AP courses which are directly comparable to our work and also consider other student assessments like Graduate Record Examinations (GREs). However, these evaluations suffer from the same problems with the scientific rigour as the AP classes.

\subsection*{Scientific assessment of ChatGPT}

Researchers across the globe are currently assessing the individual capabilities of these models with greater scientific rigour. We note that due to the recency and speed of these developments, the hereafter discussed literature has mostly only been published as pre-prints and has not yet been peer-reviewed. In addition to the above issues concretely related to the assessment of the capabilities to generate student essays, it is also worth noting that there are likely large problems with the trustworthiness of evaluations, because of data contamination, i.e., because the benchmark tasks are part of the training of the model, which enables memorization. For example, Aiyappa et al.~\cite{aiyappa2023trust} find evidence that this is likely the case for benchmark results regarding NLP tasks. This complicates the effort by researchers to assess the capabilities of the models beyond memorization. 

Nevertheless, the first assessment results are already available -- though mostly focused on ChatGPT-3 and not yet ChatGPT-4. Closest to our work is a study by Yeadon et al.~\cite{yeadon2022death}, who also investigate ChatGPT-3 performance when writing essays. They grade essays generated by ChatGPT-3 for five physics questions based on criteria that cover academic content, appreciation of the underlying physics, grasp of subject material, addressing the topic, and writing style. For each question, ten essays were generated and rated independently by five researchers. While the sample size precludes a statistical assessment, the results demonstrate that the AI model is capable of writing high-quality physics essays, but that the quality varies in a manner similar to human-written essays.

Guo et al.~\cite{guo2023close} create a set of free-text question answering tasks based on data they collected from the internet, e.g., question answering from Reddit. The authors then sample thirty triplets of a question, a human answer, and a ChatGPT-3 generated answer and ask human raters to assess if they can detect which was written by a human, and which was written by an AI. While this approach does not directly assess the quality of the output, it serves as a Turing test~\cite{turing1950} designed to evaluate whether humans can distinguish between human- and AI-produced output. The results indicate that humans are in fact able to distinguish between the outputs when presented with a pair of answers. Humans familiar with ChatGPT are also able to identify over 80\% of AI-generated answers without seeing a human answer in comparison. However, humans that are not yet familiar with ChatGPT-3 are not capable identify AI-written answers about 50\% of the time. Moreover, the authors also find that the AI-generated outputs are deemed to be more helpful than the human answers in slightly more than half of the cases. This suggests that the strong results from the owner's benchmarks regarding the capabilities to generate free-text answers generalize beyond the benchmarks. 

There are, however, some indicators that the benchmarks may be overly optimistic in their assessment of the model's capabilities. For example, Kortemeyer~\cite{kortemeyer2023artificialintelligence} conducts a case study to assess how well ChatGPT-3 would perform in a physics class, simulating the tasks that students need to complete as part of the course: answer multiple-choice questions, do homework assignments, ask questions during a lesson, complete programming exercises, and write exams with free-text questions. Notably, ChatGPT-3 was allowed to interact with the instructor for many of the tasks, allowing for multiple attempts as well as feedback on preliminary solutions. The experiment shows that ChatGPT-3's performance is in many aspects similar to that of the beginning learners and that the model makes similar mistakes, e.g., omits units, or simply plugs in results from equations. Overall, the AI would have passed the course with a low score of 1.5 out of 4.0. Similarly, Kung et al.~\cite{kung2023usmle} study the performance of ChatGPT-3 in the United States Medical Licensing Exam (USMLE) and find that the model performs at or near the passing threshold. Their assessment is a bit more optimistic than Kortemeyer's as they state that this level of performance, comprehensible reasoning, and valid clinical insights suggest that models such as ChatGPT may potentially assist human learning in clinical decision-making.

Frieder et al.~\cite{frieder2023mathematical} evaluate the capabilities of ChatGPT-3 in solving graduate-level mathematical tasks. They find that while ChatGPT-3 seems to have some mathematical understanding, its level is well below that of an average student and in most cases is not sufficient to pass exams. Yuan et al.~\cite{yuan2023large} consider the arithmetic abilities of language models, including ChatGPT-3 and ChatGPT-4. They find that they exhibit the best performance among other currently available language models (incl. Llama~\cite{touvron2023llama}, FLAN-T5~\cite{chung2022scaling}, and Bloom~\cite{workshop2023bloom}). However, the accuracy of basic arithmetic tasks is still only at 83\% when considering correctness to the degree of $10^{-3}$, i.e., such models are still not capable of functioning reliably as calculators. In a slightly satiric, yet insightful take, Spencer et al.~\cite{spencer2023ai} assess how a scientific paper on gamma-ray astrophysics would look like, if it were written largely with the assistance of ChatGPT-3. They find that while the language capabilities are good and the model is capable of generating equations, the arguments are often flawed and the references to scientific literature are full of hallucinations. 

The general reasoning skills of the models may also not be on the level expected from the benchmarks. For example, Cherian et al.~\cite{cherian2023deep} evaluate how well ChatGPT-3 performs on eleven puzzles that second-graders should be able to solve and find that ChatGPT is only able to solve them on average in 36.4\% of attempts, whereas the second graders achieve a mean of 60.4\%. However, their sample size is very small and the problem was posed as a multiple-choice question answering problem, which cannot be directly compared to the NLG we consider. 

\subsection*{Research gap}

Within this article, we address an important part of the current research gap regarding the capabilities of ChatGPT (and similar technologies), guided by the following research questions: 

\begin{description}
\item[\textbf{RQ1:}] How good is ChatGPT based on GPT-3 and GPT-4 at writing argumentative student essays?
\item[\textbf{RQ2:}] How do AI-generated essays compare to essays written by humans?
\item[\textbf{RQ3:}] What are linguistic devices that are characteristic of human versus AI-generated content?
\end{description}

We study these aspects with the help of a large group of teaching professionals that systematically assess a large corpus of essays. To the best of our knowledge, this is the first large-scale, independent scientific assessment of ChatGPT (or similar models) of this kind. The answer to these questions is vital for understanding of the impact of ChatGPT on the future of education. 

\section*{Materials and Methods}


\subsection*{Data}

The essay topics originate from a corpus of argumentative essays in the field of argument mining \cite{stab-gurevych-2014-annotating}. Argumentative essays require students to think critically about a topic and use evidence to establish a position on the topic in a concise manner. The corpus features essays for 90 topics from \url{https://essayforum.com/}, an active community for providing writing feedback on different kinds of text and is frequented by high-school students to get feedback from native speakers on their essay-writing capabilities. Topics range from `Should students be taught to cooperate or to compete?' to `Will newspapers become a thing of the past?'. In the corpus, each topic features one human-written essay uploaded and discussed in the forum. The students who wrote the essays are not native speakers. The average length of these essays is 19 sentences with 388 tokens (an average of 2.089 characters) and will be termed `student essays' in the remainder of the paper.

For the present study, we use the topics from Stab and Gurevych \cite{stab-gurevych-2014-annotating} and prompt ChatGPT with `Write an essay with about 200 words on ``[{\tt topic}]''' to receive automatically-generated essays from the ChatGPT-3 and ChatGPT-4 versions from 22 March 2023 (`ChatGPT-3 essays', `ChatGPT-4 essays'). No additional prompts for getting the responses were used, i.e., the data was created with a basic prompt in a zero-shot scenario. This is in contrast to the benchmarks by OpenAI, who used an engineered prompt in a few-shot scenario to guide the generation of essays. We note that we decided to ask for 200 words, because we noticed a tendency to generate essays that are longer than the desired length by ChatGPT. Similar to the evaluations of free-text answers by OpenAI, we did not consider multiple configurations of the model due to the effort required to obtain human judgements. For the same reason, our data is restricted to ChatGPT and does not include other models available at that time, e.g., Alpaca. We use the browser versions of the tools, because we consider this to be a more realistic scenario than using the API. Table \ref{tab:overview-stats} below shows the core statistics of the resulting dataset. 

\begin{table}[h]
    \centering
    \caption{Core statistics of the dataset}
    \begin{tabular}{l|l|l|l}
      \hline
      {\bf Source} & {\bf length (words/essay)} &  {\bf sentences/essay} & {\bf words/sentence}\\
      \hline
      Student  & 339.13 & 18.98 & 18.60\\ \hline
      ChatGPT-3 & 247.96 & 12.40 & 20.31\\ \hline
      ChatGPT-4 & 253.70 & 13.08 & 19.57\\
      \hline
    \end{tabular}
    \label{tab:overview-stats}
\end{table}

\subsection*{Annotation study}

\subsubsection*{Study participants}

The participants had registered for a two-hour online training entitled `ChatGPT -- Challenges and Opportunities' conducted by the authors of this paper as a means to provide teachers with some of the technological background of NLG systems in general and ChatGPT in particular. Only teachers permanently employed at secondary schools were allowed to register for this training. Focusing on these experts alone allows us to receive meaningful results as those participants have a wide range of experience in assessing students' writing.  
A total of 139 teachers registered for the training, 129 of them teach at grammar schools, and only 10 teachers hold a position at other secondary schools. About half of the registered teachers (68 teachers) have been in service for many years and have successfully applied for promotion. For data protection reasons, we do not know the subject combinations of the registered teachers. We only know that a variety of subjects are represented, including languages (English, French and German), religion/ethics, and science. 

The training began with an online lecture followed by a discussion phase. Teachers were given an overview of language models and basic information on how ChatGPT was developed. After about 45 minutes, the teachers received an explanation of the questionnaire at the core of our study (see Supplementary material S3) and were informed that they had 30 minutes to finish the study tasks. The explanation included information on how the data was obtained, why we collect the self-assessment, and how we chose the criteria for the rating of the essays. Participation in the questionnaire was voluntary and did not affect the awarding of a training certificate. Once these instructions were provided, the link to the online form was given to the participants. After the questionnaire, we continued with an online lecture on the opportunities for using ChatGPT for teaching as well as AI beyond chatbots.

\subsubsection*{Questionnaire}

The questionnaire consists of three parts: first, a brief self-assessment regarding the English skills of the participants which is based on the Common European Framework of Reference for Languages (CEFR).\footnote{\url{https://www.coe.int/en/web/common-european-framework-reference-languages}} We have six levels ranging from `comparable to a native speaker' to `some basic skills' (see Table~\ref{tbl:language-skills} in Supplementary material S3). Then each participant was shown six essays. The participants were only shown the generated text and were not provided with information on whether the text was human-written or AI-generated. 

The questionnaire covers the seven categories relevant for essay assessment shown below (for details see Tables~\ref{tbl:rating-categories-1} and \ref{tbl:rating-categories-2} in Supplementary material S3 at the end):  

\begin{itemize}
    \item Topic and completeness
    \item Logic and composition
    \item Expressiveness and comprehensiveness
    \item Language mastery
    \item Complexity
    \item Vocabulary and text linking
    \item Language constructs
\end{itemize}

These categories are used as guidelines for essay assessment established by the Ministry for Education of Lower Saxony, Germany.\footnote{\url{http://www.kmk-format.de/material/Fremdsprachen/5-3-2_Bewertungsskalen_Schreiben.pdf}} For each criterion, a seven-point Likert scale with scores from zero to six is defined, where zero is the worst score (e.g., no relation to the topic) and six is the best score (e.g., addressed the topic to a special degree). The questionnaire included a written description as guidance for the scoring. 

After rating each essay, the participants were also asked to self-assess their confidence in the ratings. We used a five-point Likert scale based on the criteria for the self-assessment of peer-review scores from the Association for Computational Linguistics (ACL). Once a participant finished rating the six essays, they were shown a summary of their ratings, as well as the individual ratings for each of their essays and the information on how the essay was generated. 

\subsection*{Computational linguistic analysis}

In order to further explore and compare the quality of the essays written by students and ChatGPT, we consider the six following linguistic characteristics: lexical diversity, sentence complexity, nominalisation, presence of modals, epistemic and discourse markers. Those are motivated by previous work: Weiss et al. \cite{weiss-etal-2019-computationally} observe the correlation between measures of lexical, syntactic and discourse complexities to the essay gradings of German high-school examinations while McNamara
et al. \cite{mcnamara2010linguistic} explore cohesion (indicated, among other things, by connectives), syntactic complexity and lexical diversity in relation to the essay scoring.

\subsubsection*{Lexical diversity}
We identify vocabulary richness by using a well-established measure of textual, lexical diversity (MTLD) \cite{mccarthy2010mtld} which is often used in the field of automated essay grading \cite{weiss-etal-2019-computationally, mcnamara2010linguistic, dasgupta2018augmenting}. It takes into account the number of unique words but unlike the best-known measure of lexical diversity, the type-token ratio (TTR), it is not as sensitive to the difference in the length of the texts. In fact, Koizumi and In’nami \cite{koizumi2012effects} find it to be least affected by the differences in the length of the texts compared to some other measures of lexical diversity. This is relevant to us due to the difference in average length between the human-written and ChatGPT-generated essays. 

\subsubsection*{Syntactic complexity}
We use two measures in order to evaluate the syntactic complexity of the essays. One is based on the maximum depth of the sentence dependency tree which is produced using the spaCy 3.4.2 dependency parser \cite{spacy} (`Syntactic complexity (depth)'). For the second measure we adopt an approach similar in nature to the one by Weiss et al.~ \cite{weiss-etal-2019-computationally}  who use clause structure to evaluate syntactic complexity. In our case, we count the number of conjuncts, clausal modifiers of nouns, adverbial clause modifiers, clausal complements, clausal subjects, and parataxes (`Syntactic complexity (clauses)'). The supplementary material in S2 shows the difference between sentence complexity based on two examples from the data.

Nominalisation is a common feature of a more scientific style of writing \cite{siskou2022measuring} and is used as an additional measure for syntactic complexity. In order to explore this feature, we count occurrences of nouns with suffixes such as `-ion', `-ment', `-ance' and a few others which are known to transform verbs into nouns. 

\subsubsection*{Semantic properties}
Both modals and epistemic markers signal the commitment of the writer to their statement. We identify modals using the POS-tagging module provided by spaCy as well as a list of epistemic expressions of modality, such as `definitely' 
and `potentially', also used in other approaches to identifying semantic properties \cite{2020elassadydiscourse}. For epistemic markers we adopt an empirically-driven approach and utilize the epistemic markers identified in a corpus of dialogical argumentation by Hautli-Janisz et al.~\cite{hautli-janisz-etal-2022-qt30}. We consider expressions such as `I think', `it is believed' and `in my opinion' to be epistemic.

\subsubsection*{Discourse properties}
Discourse markers can be used to measure the coherence quality of a text. This has been explored by Somasundaran et al. \cite{somasundaran2018towards} who use discourse markers to evaluate the story-telling aspect of student writing while Nadeem et al.~\cite{nadeem-etal-2019-automated} incorporated them in their deep learning-based approach to automated essay scoring. In the present paper, we employ the PDTB list of discourse markers \cite{prasad-etal-2008-penn} which we adjust to exclude words that are often used for purposes other than indicating discourse relations, such as `like', `for', `in' etc.

\subsection*{Statistical methods}

We use a within-subjects design for our study. Each participant was shown six randomly selected essays. Results were submitted to the survey system after each essay was completed, in case participants ran out of time and did not finish scoring all six essays. Cronbach's $\alpha$ allows us to determine the inter-rater reliability for the rating criterion and data source (human, ChatGPT-3, ChatGPT-4) in order to understand the reliability of our data not only overall, but also for each data source and rating criterion. We use two-sided Wilcoxon-ranksum tests to confirm the significance of the differences between the data sources for each criterion. We use the same tests to determine the significance of the linguistic characteristics. This results in three comparisons (human vs. ChatGPT-3, human vs. ChatGPT-4, ChatGPT-3 vs. ChatGPT-4) for each of the seven rating criteria and each of the seven linguistic characteristics, i.e., 42 tests. We use the Holm-Bonferroni method for the correction for multiple tests to achieve a family-wise error rate of 0.05. We report the effect size using Cohen's $d$. While our data is not perfectly normal, it also does not have severe outliers, so we prefer the clear interpretation of Cohen's $d$ over the slightly more appropriate, but less accessible non-parametric effect size measures. We report point plots with estimates of the mean scores for each data source and criterion, incl.~the 95\% confidence interval of these mean values. The confidence intervals are estimated in a non-parametric manner based on bootstrap sampling. We further visualize the distribution for each criterion using violin plots to provide a visual indicator of the spread of the data (see S4 in Supplementary material).

Further, we use the self-assessment of the English skills and confidence in the essay ratings as confounding variables. Through this, we determine if ratings are affected by the language skills or confidence, instead of the actual quality of the essays. We control for the impact of these by measuring Pearson's rank correlation $r$ between the self-assessments and the ratings. We also determine whether the linguistic features are correlated with the ratings as expected. The sentence complexity (both tree depth and dependency clauses), as well as the nominalization, are indicators of the complexity of the language.  Similarly, the use of discourse markers should signal a proper logical structure. Finally, a large lexical diversity should be correlated with the ratings for the vocabulary. Same as above, we measure Pearson's $r$. 

Our statistical analysis of the data is implemented in Python. We use pandas 1.5.3 and numpy 1.24.2 for the processing of data and calculation of Pearson's $r$, pingouin 0.5.3 for the calculation of Cronbach's $\alpha$, scipy 1.10.1 for the Wilcoxon-ranksum-tests, and seaborn 0.12.2 for the generation of plots, incl.~the calculation of error bars that visualize the confidence intervals. 

\subsubsection*{Replication package}

All materials are available online in form of a replication that contains the data and the analysis code.\footnote{https://github.com/sherbold/chatgpt-student-essay-study}

\section*{Results}

\begin{table}
\centering
\small
\caption{Arithmetic mean (M), standard deviation (SD), and Cronbach's $\alpha$ for the ratings.}
\begin{tabular}{l|c|c|c|c|c|c|c|c|c}
\hline
 &  \multicolumn{3}{c|}{\textbf{Humans}} & \multicolumn{3}{c|}{\textbf{ChatGPT-3}} & \multicolumn{3}{c}{\textbf{ChatGPT-4}} \\ \hline
 \textbf{Criterion} & M & SD & $\alpha$ & M & SD & $\alpha$ & M & SD & $\alpha$ \\
\hline
Topic and completeness & 3.58 & 1.30 & 0.95 & 4.24 & 1.16 & 0.95 & 4.54 & 1.12 & 0.95 \\ \hline
Logic and composition & 3.64 & 1.27 & 0.96 & 4.29 & 1.04 & 0.96 & 4.64 & 1.01 & 0.96 \\ \hline
Expressiveness and compr. & 3.42 & 1.25 & 0.95 & 3.90 & 1.04 & 0.95 & 4.23 & 1.12 & 0.95 \\ \hline
Language mastery & 3.90 & 1.37 & 0.89 & 5.03 & 1.19 & 0.89 & 5.25 & 1.08 & 0.89 \\ \hline
Complexity & 3.72 & 1.26 & 0.92 & 4.20 & 1.14 & 0.92 & 4.60 & 1.10 & 0.92 \\ \hline
Vocabulary and text linking & 3.78 & 1.18 & 0.97 & 4.41 & 1.05 & 0.97 & 4.81 & 1.06 & 0.97 \\ \hline
Language constructs & 3.80 & 1.15 & 0.97 & 4.47 & 1.02 & 0.97 & 4.73 & 1.07 & 0.97 \\
\hline
Overall & 3.69 & 1.26 & & 4.36 & 1.14 & & 4.68 & 1.11 & \\
\hline
\end{tabular}
\label{tbl:means}
\end{table}

\begin{table}
\small
\caption{Arithmetic mean (M) and standard deviation (SD) for the linguistic markers.}
\begin{tabular}{l|c|c|c|c|c|c}
\hline
 \textbf{Linguistic characteristic} &  \multicolumn{2}{c|}{\textbf{Humans}} & \multicolumn{2}{c|}{\textbf{ChatGPT-3}} & \multicolumn{2}{c}{\textbf{ChatGPT-4}} \\
 \hline
 Lexical diversity & 95.72 & 23.50 & 75.68 & 12.89 & 108.91 & 20.73 \\ \hline
Syntactic complexity (depth) & 5.72 & 0.80 & 6.18 & 0.76 & 5.94 & 0.54 \\ \hline
Syntactic complexity (clauses) & 1.81 & 0.57 & 2.31 & 0.50 & 2.08 & 0.42 \\ \hline
Nominalizations & 1.06 & 0.51 & 1.56 & 0.63 & 1.73 & 0.49 \\ \hline
Modals & 10.84 & 5.30 & 8.97 & 4.21 & 6.12 & 3.18 \\ \hline
Epistemic markers & 0.06 & 0.06 & 0.02 & 0.03 & 0.00 & 0.00 \\ \hline
Discourse markers & 0.57 & 0.24 & 0.52 & 0.19 & 0.36 & 0.17 \\ \hline
\end{tabular}
\label{tbl:means-linguistic}
\end{table}

\begin{table}[]
\centering
\small
\caption{P-values of the Wilcoxon signed-rank tests adjusted for multiple comparisons using the Holm-Bonferroni method. Effect sizes measured with Cohen's $d$ reported for significant results.}
\begin{tabular}{l|r|r|r}
\hline
& \textbf{Human} & \textbf{Human} & \textbf{ChatGPT-3} \\
\textbf{Criterion /} & \textbf{vs.} & \textbf{vs.} & \textbf{vs.} \\
\textbf{Linguistic characteristic} & \textbf{ChatGPT-3} & \textbf{ChatGPT-4} & \textbf{ChatGPT-4} \\
\hline
Topic and completeness & $<$0.001 (d=-0.77) & $<$0.001 (d=-1.09) & 0.095 \\ \hline
Logic and composition & $<$0.001 (d=-0.84) & $<$0.001 (d=-1.20) & 0.025 (d=-0.45) \\ \hline
Expressiveness and compr. & 0.008 (d=-0.57) & $<$0.001 (d=-0.88) & 0.055 \\ \hline
Language mastery & $<$0.001 (d=-1.15) & $<$0.001 (d=-1.43) & 0.105 \\ \hline
Complexity & 0.025 (d=-0.52) & $<$0.001 (d=-0.99) & 0.025 (d=-0.48) \\ \hline
Vocabulary and text linking & $<$0.001 (d=-0.76) & $<$0.001 (d=-1.27) & 0.012 (d=-0.50) \\ \hline
Language constructs & $<$0.001 (d=-0.82) & $<$0.001 (d=-1.15) & 0.105 \\
\thickhline
Lexical diversity & $<$0.001 (d=1.06) & 0.001 (d=-0.60) & $<$0.001 (d=-1.93) \\ \hline
Syntactic complexity (depth) & 0.001 (d=-0.59) & 0.055 & 0.105 \\ \hline
Syntactic complexity (clauses) & $<$0.001 (d=-0.93) & 0.004 (d=-0.54) & 0.024 (d=0.49) \\ \hline
Nominalizations & $<$0.001 (d=-0.88) & $<$0.001 (d=-1.35) & 0.020 (d=-0.29) \\ \hline
Modals & 0.025 (d=0.39) & $<$0.001 (d=1.08) & $<$0.001 (d=0.76) \\ \hline
Epistemic markers & $<$0.001 (d=1.01) & $<$0.001 (d=1.53) & 0.005 (d=0.65) \\ \hline
Discourse markers & 0.150 & $<$0.001 (d=0.98) & $<$0.001 (d=0.85) \\ \hline
\end{tabular}
\label{tbl:pvals}
\end{table}

Out of the 111 teachers that completed the questionnaire, 108 rated all six essays, one rated five essays, one rated two essays, and one rated only one essay. This results in 658 ratings for 270 essays (90 topics for each essay type: human-, ChatGPT-3-, ChatGPT-4-generated), with three ratings for 121 essays, two ratings for 144 essays, and one rating for five essays. The inter-rater agreement is consistently excellent ($\alpha>0.9$), with the exception of language mastery where we have good agreement ($\alpha=0.89$, see Table~\ref{tbl:means}). Further, the correlation analysis depicted in Fig~\ref{fig:correlations} in Supplementary material S4 shows that neither the self-assessment for the English skills nor the self-assessment for the confidence in ratings is correlated with the actual ratings. Overall, this indicates that our ratings are reliable estimates of the actual quality of the essays. 

Table~\ref{tbl:means} and Fig~\ref{fig:result-distirbution-criteria} in Supplementary material S4 characterize the distribution of the ratings for the essays, grouped by the data source. We observe that for all criteria, we have a clear order of the mean values, with students having the worst ratings, ChatGPT-3 in the middle rank, and ChatGPT-4 with the best performance. We further observe that the standard deviations are fairly consistent and slightly larger than one, i.e., the spread is similar for all ratings and essays. This is further supported by the visual analysis of the violin plots. 

The statistical analysis of the ratings reported in Table~\ref{tbl:pvals} shows that differences between the human-written essays and the ones generated by both ChatGPT models are significant. The effect sizes for human versus ChatGPT-3 essays are between 0.52 and 1.15, i.e., a medium ($d \in [0.5,0.8)$) to large ($d \in [0.8, 1.2)$) effect. On the one hand, the smallest effects are observed for the expressiveness and complexity, i.e., when it comes to the overall comprehensiveness and complexity of the sentence structures, the differences between the humans and the ChatGPT-3 model are smallest. On the other hand, the difference in language mastery is larger than all other differences, which indicates that humans are more prone to making mistakes when writing than the NLG models. The magnitude of differences between humans and ChatGPT-4 is larger with effect sizes between 0.88 and 1.43, i.e., a large to very large ($d \in [1.2, 2)$) effect. Same as for ChatGPT-3, the differences are smallest for expressiveness and complexity and largest for language mastery. Please note that the difference in language mastery between humans and both GPT models does not mean that the humans have low scores for language mastery (M=3.90), but rather that the NLG models have exceptionally high scores (M=5.03 for ChatGPT-3, M=5.25 for ChatGPT-4).

When we consider the differences between the two GPT models, we observe that while ChatGPT-4 has consistently higher mean values for all criteria, only the differences for logic and composition, vocabulary and text linking, and complexity are significant. The effect sizes are between 0.45 and 0.5, i.e., small ($d \in [0.2, 0.5)$) and medium. Thus, while GPT-4 seems to be an improvement over GPT-3.5 in general, the only clear indicator of this is a better and clearer logical composition, and more complex writing with a more diverse vocabulary. 

We also observe significant differences in the distribution of linguistic characteristics between all three groups. Sentence complexity (depth) is the only category without significant difference between humans and ChatGPT-3, as well as ChatGPT-3 and ChatGPT-4. There is also no significant difference in the category of discourse markers between humans and ChatGPT-3. The magnitude of the effects varies a lot and is between 0.39 and 1.93, i.e., between small ($d \in [0.2, 0.5)$) and very large. However, in comparison to the ratings, there is no clear tendency regarding the direction of the differences. For instance, while the ChatGPT models write more complex sentences and use more nominalizations, humans tend to use more modals and epistemic markers instead. The lexical diversity of humans is higher than that of ChatGPT-3 but lower than that of ChatGPT-4. While there is no difference in the use of discourse markers between humans and ChatGPT-3, ChatGPT-4 uses significantly fewer discourse markers. 

We detect the expected positive correlations between the complexity ratings and the linguistic markers for sentence complexity ($r=0.16$ for depth, $r=0.19$ for clauses) and nominalizations ($r=0.22$). However, we observe a negative correlation between the logic ratings and the discourse markers ($r=-0.14$), which counters our intuition that more frequent use of discourse indicators makes a text more logically coherent. However, this is in line with previous work: McNamara
et al.\cite{mcnamara2010linguistic} also find no indication that the use of cohesion indices such as discourse connectives correlates with high- and low-proficiency essays. Finally, we observe the expected positive correlation between the ratings for the vocabulary and the lexical diversity ($r=0.12$). We note that the strength of all these correlations is weak.

\section*{Discussion}

Our results provide clear answers to the first two research questions that consider the quality of the generated essays: ChatGPT performs well at writing argumentative student essays and outperforms humans significantly. The ChatGPT-4 model has (at least) a large effect and is on average about one point better than humans on a seven-point Likert scale. 

Regarding the third research question, we find that there are significant linguistic differences between humans and AI-generated content. The AI-generated essays are highly structured, which for instance is reflected by the identical beginnings of the concluding sections of all ChatGPT essays (`In conclusion, [...]'). The initial sentences of each essay are also very similar starting with a general statement using the main concepts of the essay topics. Although this corresponds to the general structure that is sought after for argumentative essays, it is striking to see that the ChatGPT models are so rigid in realizing this, whereas the human-written essays are looser in representing the guideline on the linguistic surface. Moreover, the linguistic fingerprint has the counter-intuitive property that the use of discourse markers is negatively correlated with logical coherence. We believe that this might be due to the rigid structure of the generated essays: instead of using discourse markers, the AI models provide a clear logical structure by separating the different arguments into paragraphs, thereby reducing the need for discourse markers.

Our data also shows that hallucinations are not a problem in the setting of argumentative essay writing: the essay topics are not really about factual correctness, but rather about argumentation and critical reflection on general concepts which seem to be contained within the knowledge of the AI model. The stochastic nature of the language generation is well-suited for this kind of task, as different plausible arguments can be seen as a sampling from all available arguments for a topic. Nevertheless, we need to perform a more systematic study of the argumentative structures in order to better understand the difference in argumentation between human-written and ChatGPT-generated essay content.

One of the issues  with evaluations of the recent large-language models is not accounting for the impact of tainted data when benchmarking such models. While it is certainly possible that the essays that were sourced by Stab and Gurevych~\cite{stab-gurevych-2014-annotating} from the internet were part of the training data of the GPT models, the proprietary nature of the model training means that we cannot confirm this. However, we note that the generated essays did not resemble the corpus of human essays at all. Moreover, the topics of the essays are general in the sense that any human should be able to reason and write about these topics, just by understanding concepts like `cooperation'. Consequently, a taint on these general topics, i.e., the fact that they might be present in the data, is not only possible but is actually expected and unproblematic, as it relates to the capability of the models to learn about concepts, rather than the memorization of specific task solutions. 

While we did everything to ensure a sound construct and a high validity of our study, there are still certain issues which may affect our conclusions. Most importantly, neither the writers of the essays, nor their raters, were English native speakers. However, the students purposefully used a forum for English writing frequented by native speakers to ensure the language and content quality of their essays. The teachers were informed that part of the training would be in English to prevent registrations from people without English language skills. Moreover, the self-assessment of the language skills was not correlated with the ratings, indicating that the threat to the soundness of our results is low. While we cannot definitively rule out that our results would not be reproducible with other human raters, the high inter-rater agreement indicates that this is unlikely. We further note that the essay topics may not be an unbiased sample. While Stab and Gurevych~\cite{stab-gurevych-2014-annotating} randomly sampled the essays from the writing feedback section of an essay forum, it is unclear whether the essays posted there are representative of the general population of essay topics. Nevertheless, we believe that the threat is fairly low, because our results are consistent and do not seem to be influenced by certain topics. Further, we cannot with certainty conclude how our results generalize beyond ChatGPT-3 and ChatGPT-4 to similar models like Bard,\footnote{\url{https://bard.google.com/?hl=en}} Alpaca, and Dolly. Especially the results for linguistic characteristics are hard to predict. However, since -- to the best of our knowledge and given the proprietary nature of some of these models -- the general approach to how these models work is similar and the trends for essay quality should hold for models with comparable size and training procedures. 

\section*{Conclusion}

Our results provide a strong indication that the fear many teaching professionals have is warranted: the way students do homework and teachers assess it needs to change in a world of generative AI models. Our results show that when students want to maximize their essay grades, they could easily do so by relying on results from AI models like ChatGPT. However, this is not and cannot be the goal of education. Consequently, educators need to change how they approach homework. Instead of just assigning and grading essays, we need to reflect more on the output of AI tools regarding their reasoning and correctness. AI models need to be seen as an integral part of education, but one which requires careful reflection and training of critical thinking skills. 

Furthermore, teachers need to adapt strategies for teaching writing skills: 
as with the use of calculators, it is necessary to critically reflect with the students on when and how to use those tools.  For instance, constructivists \cite{windschitl2002constructivism} argue that learning is enhanced by the active design and creation of unique artefacts by students themselves. In the present case this means that, in the long term, educational objectives may need to be adjusted. This is analogous to teaching good arithmetic skills to younger students and then allowing and encouraging students to use calculators freely in later stages of education. Similarly, once a sound level of literacy has been achieved, strongly integrating AI models in lesson plans may no longer run counter to reasonable learning goals. Instead, it can be seen as an opportunity to redesign competency-based learning objectives, for example, to focus on the ability to analyse and rate the logical structure, cohesion and coherence of (generated) texts. More generally, teachers and researchers should embrace the invitation to explore, understand, evaluate, and assess a wide range of AI contributions to and impacts on the teaching/learning process as a whole.

In terms of shedding light on the quality and structure of AI-generated essays, this paper makes an important contribution by offering an independent, large-scale and statistically sound account of essay quality, comparing human-written and AI-generated texts. By comparing different versions of ChatGPT, we also offer a glance into the development of these models over time in terms of their linguistic properties and the quality they exhibit.

\section*{Supporting information}

\paragraph*{S1 Essay examples from the dataset}
\label{S1_Fig}
 
 Three essays (student essay, ChatGPT-3 generated essay and ChatGPT-4 generated essay) for the topic `Should students be taught to compete or to cooperate?'
\begin{quote}
    
{\bf Student essay: }It is always said that competition can effectively promote the development of economy. In order to survive in the competition, companies continue to improve their products and service, and as a result, the whole society prospers. However, when we discuss the issue of competition or cooperation, what we are concerned about is not the whole society, but the development of an individual's whole life. From this point of view, I firmly believe that we should attach more importance to cooperation during primary education.
 
 First of all, through cooperation, children can learn about interpersonal skills which are significant in the future life of all students. What we acquired from team work is not only how to achieve the same goal with others but more importantly, how to get along with others. During the process of cooperation, children can learn about how to listen to opinions of others, how to communicate with others, how to think comprehensively, and even how to compromise with other team members when conflicts occurred. All of these skills help them to get on well with other people and will benefit them for the whole life.
 
 On the other hand, the significance of competition is that how to become more excellence to gain the victory. Hence it is always said that competition makes the society more effective. However, when we consider about the question that how to win the game, we always find that we need the cooperation. The greater our goal is, the more competition we need. Take Olympic games which is a form of competition for instance, it is hard to imagine how an athlete could win the game without the training of his or her coach, and the help of other professional staffs such as the people who take care of his diet, and those who are in charge of the medical care. The winner is the athlete but the success belongs to the whole team. Therefore without the cooperation, there would be no victory of competition.
 
 Consequently, no matter from the view of individual development or the relationship between competition and cooperation we can receive the same conclusion that a more cooperative attitudes towards life is more profitable in one's success.
 \end{quote}
 
 \begin{quote}
     
{\bf ChatGPT-3 essay: }Education is not only about acquiring knowledge, but also about developing the skills and attitudes necessary to succeed in life. One of the most important decisions in education is whether students should be taught to compete or to cooperate.

On the one hand, competition is often seen as a way to motivate students to work harder and achieve more. By pitting students against one another, it is believed that they will be pushed to excel and develop important skills such as resilience, determination, and strategic thinking. In a competitive environment, students may also learn to be independent and self-reliant, which are valuable traits in the real world.

On the other hand, cooperation is equally important in preparing students for their future. By working together, students learn to communicate, share ideas, and build relationships. Cooperation also helps to promote empathy and respect for others, as students learn to appreciate the strengths and weaknesses of their peers. Furthermore, in many industries and careers, collaboration and teamwork are essential for success, and students who are not taught to cooperate may struggle to adapt to these environments.

In conclusion, both competition and cooperation have their advantages and disadvantages, and both are important skills for students to develop. However, it is essential that educators strike a balance between the two, teaching students when to compete and when to cooperate, and providing opportunities to practice both skills. Ultimately, the goal should be to prepare students for the diverse and complex challenges they will face in the future, whether they arise from competition or cooperation.
 \end{quote}

 \begin{quote}
{\bf ChatGPT-4 essay: }The contemporary education system frequently grapples with the question of whether students should be taught to compete or cooperate. Striking the right balance is essential, as both competition and cooperation have their merits in shaping a well-rounded individual.

Competition can act as a powerful motivator, pushing students to excel and surpass their perceived limitations. It fosters a spirit of resilience, adaptability, and innovation, as learners strive to outperform their peers. Moreover, competition prepares students for real-world challenges, where they are bound to encounter similar competitive scenarios in their professional lives.

However, solely focusing on competition can lead to excessive stress, anxiety, and unhealthy rivalry, which may hinder students' personal and academic growth. This is where the value of cooperation becomes apparent. By teaching students to cooperate, educators encourage essential skills like teamwork, communication, empathy, and problem-solving. These skills are indispensable in today's increasingly interconnected world, where success often depends on the ability to collaborate with others.

In conclusion, the ideal education system should strike a balance between competition and cooperation. While competition nurtures a drive for excellence and adaptability, cooperation fosters critical interpersonal skills and mutual support. By integrating both elements, educators can help shape well-rounded individuals who are prepared to excel in their personal and professional lives.
\end{quote}

\paragraph*{S2 Examples for language complexity}
\label{S2_Fig}

Example {\ref{ex1}} is one of the simpler sentences from the student essay from our dataset which can be found in \nameref{S1_Fig}. This sentence has zero clauses and its maximum dependency tree depth is three. The sentence in Example {\ref{ex2}} from the same student essay is significantly longer with four clauses and a maximum dependency tree depth of nine.

\enumsentence{\textit{The greater our goal is, the more competition we need.\label{ex1} }}
\enumsentence{\textit{Take Olympic games which is a form of competition for instance, it is hard to imagine how an athlete could win the game without the training of his or her coach, and the help of other professional staffs such as the people who take care of his diet, and those who are in charge of the medical care. \label{ex2} }}

\newpage 
\paragraph*{S3 Details about the variables collected within the questionnaire}

\begin{table}
\caption{Categories and levels for the rating of the essays. Continued in Table~\ref{tbl:rating-categories-2}}
\begin{tabular}{p{1.6cm}|p{8.4cm}|c}
\hline
\textbf{Category} & \textbf{Description of levels from best to worst} & \textbf{Score} \\
\hline
\multirow{13}{*}{\rotatebox[origin=c]{90}{Topic and completeness}}
& Addressed the topic to a special degree and considered all aspects of the task. & 6 \\ \cline{2-3}
& Addressed the topic in a good manner and considered all aspects of the task. & 5 \\ \cline{2-3}
& Addressed the topic and considered most aspects of the task. & 4 \\ \cline{2-3}
& Mostly addressed the topic, considered only some aspects of the task. & 3 \\ \cline{2-3}
& Topic still discernable, superficial and one-sided, of digresses and repeats itself. & 2 \\ \cline{2-3}
& Topic somewhat discernable, every superficial and one-sided. & 1 \\ \cline{2-3}
& No relation to the topic. & 0 \\
\thickhline
\multirow{10}{*}{\rotatebox[origin=c]{90}{Logic and composition}}
& Clear structure and logical and profound reasoning. & 6 \\ \cline{2-3}
& Logical and consistent structure, sound reasoning. & 5 \\ \cline{2-3}
& Mostly logical structure, sometimes chains of individual thoughts. & 4 \\ \cline{2-3}
& Errors in the logical structure, thoughts not presented coherently. & 3 \\ \cline{2-3}
& Logical structure recognizable, jumps in reasoning. & 2 \\ \cline{2-3}
& Logical structure recognizable, few thoughts brought to a conclusion. & 1 \\ \cline{2-3}
& No logical structure. & 0 \\
\thickhline
\multirow{8}{*}{\rotatebox[origin=c]{90}{\parbox{3cm}{Expressiveness and comprehensiveness}}}
& Very comprehensive and detailed depiction, consistently persuasive. & 6 \\ \cline{2-3}
& Comprehensive and detailed depiction. & 5 \\ \cline{2-3}
& Partially detailed and mostly comprehensive. & 4 \\ \cline{2-3}
& Few details, but still comprehensive. & 3 \\ \cline{2-3}
& Somewhat comprehensive. & 2 \\ \cline{2-3}
& Almost not comprehensive. & 1 \\ \cline{2-3}
& Not comprehensive. & 0 \\
\thickhline
\multirow{11}{*}{\rotatebox[origin=c]{90}{Language mastery}}
& Almost perfect language use, few minor rule violations that do not influence the understandability. & 6 \\ \cline{2-3}
& Some minor rule violations that do not have a major influence on the understandability. & 5 \\ \cline{2-3}
& Some rule violations that affect the understandability. & 4 \\ \cline{2-3}
& Multiple rule violations that affect the understandability. & 3 \\ \cline{2-3}
& Many rule violations that affect the understandability. & 2 \\ \cline{2-3}
& Many severe rule violations that affect the understandability. & 1 \\ \cline{2-3}
& Unintelligible language use. & 0 \\
\hline

\end{tabular}
\label{tbl:rating-categories-1}
\end{table}

\begin{table}
\caption{Categories and levels for the rating of the essays. Continued from Table~\ref{tbl:rating-categories-1}}
\begin{tabular}{p{1.6cm}|p{8.4cm}c}
\hline
\textbf{Category} & \textbf{Description of levels from best to worst} & \textbf{Score} \\
\hline
\multirow{10}{*}{\rotatebox[origin=c]{90}{Complexity}}
& Often uses complex sentences and sentence links. & 6 \\ \cline{2-3}
& Multiple uses complex sentences and sentence links. & 5 \\ \cline{2-3}
& Few uses of complex sentences, but multiple simple sentence links. & 4 \\ \cline{2-3}
& Consistently simple but correct sentence structures, some sentence links. & 3 \\ \cline{2-3}
& Lack of complex language, errors in sentence structures, few sentence links. & 2 \\ \cline{2-3}
& Severere errors in sentence structures, no sentence links. & 1 \\ \cline{2-3}
& No discernable sentence structures. & 0 \\
\thickhline
\multirow{13}{*}{\rotatebox[origin=c]{90}{Vocabulary and text linking}}
& Very comprehensive and variable use of vocabulary and consistent and correct linking of text elements. & 6 \\ \cline{2-3}
& Comprehensive and variable use of vocabulary, often correct linking of text elements. & 5 \\ \cline{2-3}
& Fundamental vocabulary, sometimes correct linking of text elements. & 4 \\ \cline{2-3}
& Limited vocabulary, rudimentary linking of text elements. & 3 \\ \cline{2-3}
& Limited vocabulary, sometimes bad choice of words, no linking of text elements. & 2 \\ \cline{2-3}
& Strongly limited vocabulary, often bad choice of words, no linking of text elements. & 1 \\ \cline{2-3}
& No consistent English vocabulary. & 0 \\
\thickhline
\multirow{8}{*}{\rotatebox[origin=c]{90}{Language constructs}}
& Confident use of a large number of language constructs. & 6 \\ \cline{2-3}
& Confident use of language constructs. & 5 \\ \cline{2-3}
& Mostly correct use of language constructs. & 4 \\ \cline{2-3}
& Only few language constructs and partially wrong use. & 3 \\ \cline{2-3}
& Predominantly wrong use of language constructs. & 2 \\ \cline{2-3}
& Almost always wrong use of language constructs. & 1 \\ \cline{2-3}
& Language constructs not following English language rules. & 0 \\
\hline
\end{tabular}
\label{tbl:rating-categories-2}
\end{table}

\begin{table}
\centering
\caption{Skill levels for self-assessment for reading proficiency, their CEFR equivalents, and the score on the underlying Likert scale.}
\begin{tabular}{l|c|c}
\hline
\textbf{Self-assessment} & \textbf{CEFR} & \textbf{Score} \\
\hline
Comparable to native speaker & C2 & 5\\ \hline
Business fluent& C1 & 4 \\ \hline
Fluent & B2 & 3 \\ \hline
Good & B1 & 2 \\ \hline
Good basic skills & A2 & 1 \\ \hline
Some basic skills & A1 & 0 \\ \hline
\end{tabular}
\label{tbl:language-skills}
\end{table}

\begin{table}
\centering
\caption{Scale for the self-assessment of the confidence in the essay ratings.}
\begin{tabular}{p{6.5cm}|c}
\hline
\textbf{Self-assessment of rating confidence} & \textbf{Score} \\
\hline
I am certain that my assessment is correct. & 4 \\ \hline
I am fairly certain that my assessment is correct. & 3 \\ \hline
I am fairly certain that my assessment is correct, but I may have misjudged some aspects. & 2 \\ \hline
I would defend my assessment, but it is possible that I misjudged aspects. & 1\\ \hline
I am not familiar with the topic. My assessment is based on guesses. & 0 \\ \hline
\end{tabular}
\label{tbl:confidence}
\end{table}

\FloatBarrier

\paragraph{S4 Supporting visualizations for the data analysis}

\begin{figure}
\centering
\includegraphics[width=\textwidth]{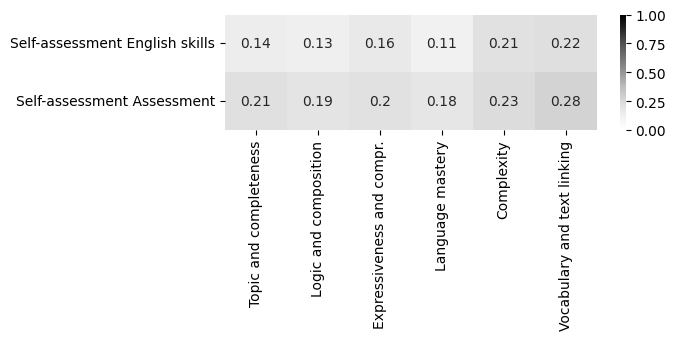}
\caption{Pearson's $r$ correlation between the self-assessments.}
\label{fig:correlations}
\end{figure}

\begin{figure}
\centering
\includegraphics[width=0.6\textwidth]{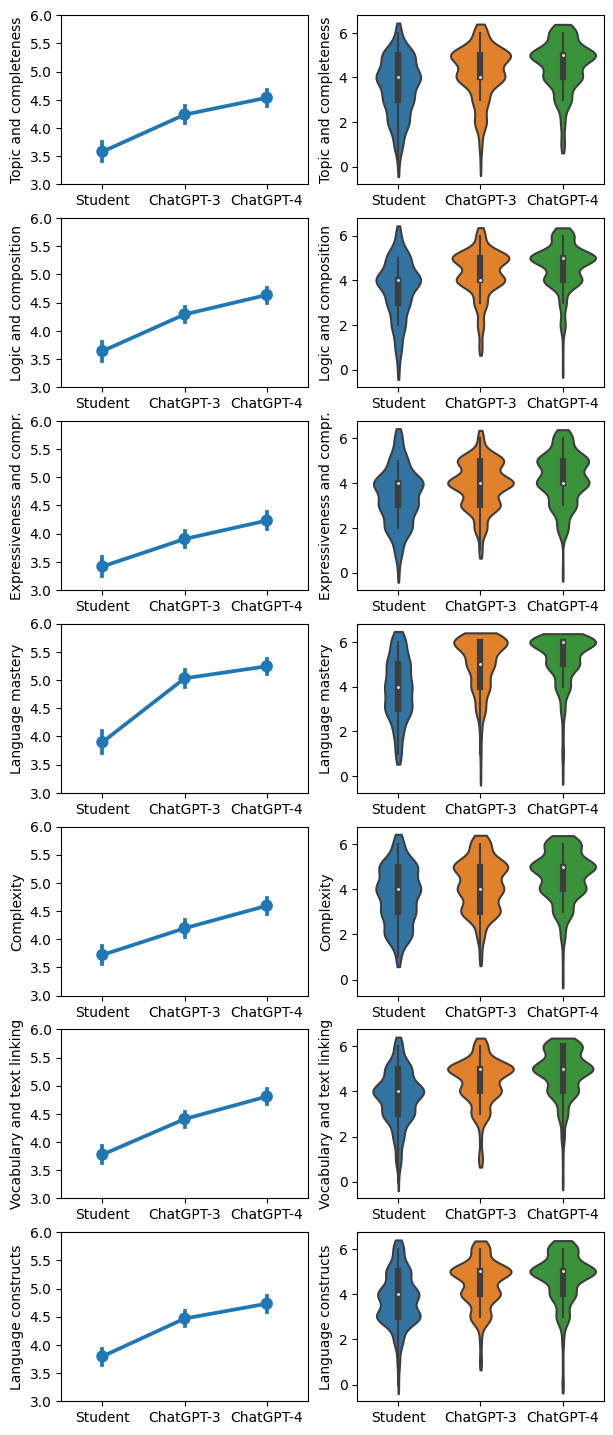}
\caption{Left: Point plots depicting the mean values and confidence intervals of the survey results. Right: Violin plots depicting the distribution of the ratings.}
\label{fig:result-distirbution-criteria}
\end{figure}

\begin{figure}
\centering
\includegraphics[width=0.6\textwidth]{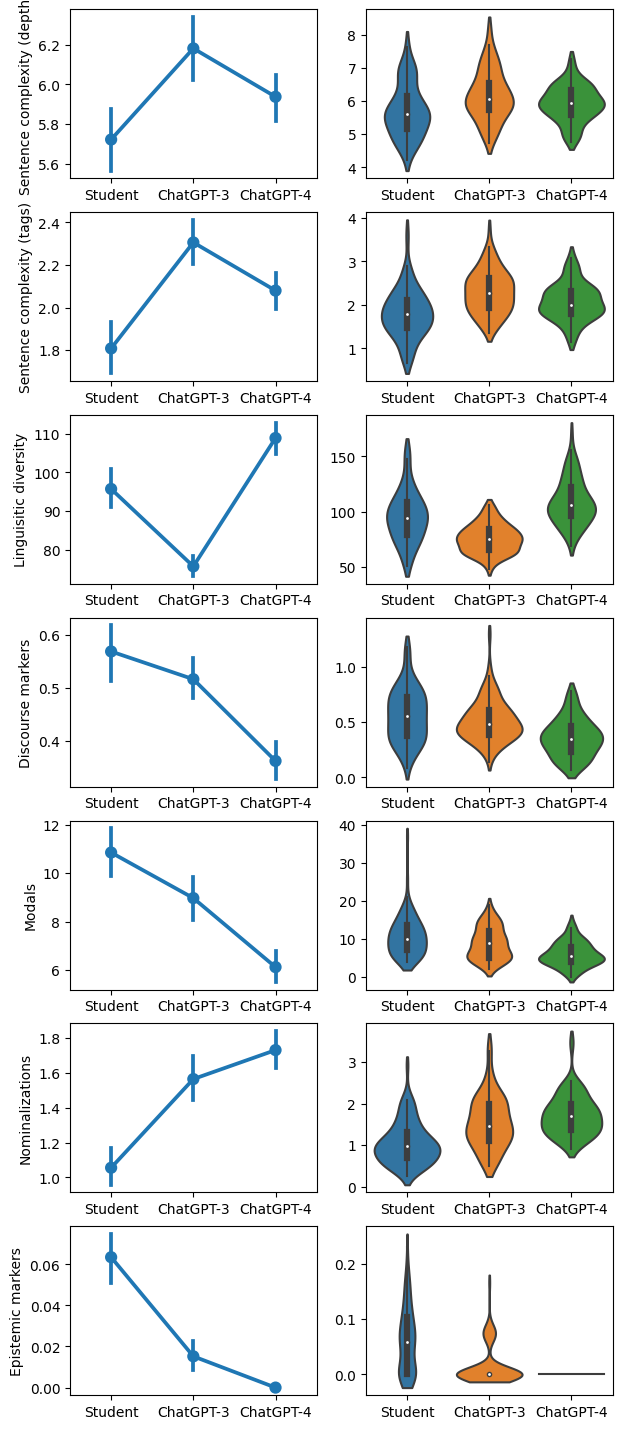}
\caption{Left: Point plots depicting the mean values and confidence intervals of the linguistic characteristics. Right: Violin plots depicting the distribution of the linguistic characteristics.}
\label{fig:result-distirbution-linguistics}
\end{figure}


\FloatBarrier
\bibliography{custom}

%
%
%
\end{document}